# Energy Variation in Training Modern Computer Vision Architectures

**David Cortes[1], Carlos Juiz[1], Belen Bermejo[1]**
[1]University of the Balearic Islands (Spain)
d.cortes@uib.es

*Abstract*—**The rapid growth of deep learning has substantially increased the energy consumption associated with model training, making energy efficiency an increasingly relevant design criterion. This study empirically measures the energy variation of training seven modern computer vision architectures—MobileNetV3-Small, MobileNetV3-Large, EfficientNet-B0, EfficientNet-B1, ViT-B/32, ConvNeXt-Tiny, and ViT-B/16—for the ImageNet-1k classification task, using a homogeneous 10,000-image subset (ImageNet-10k) and a uniform 40-epoch baseline configuration executed on two NVIDIA Tesla P100 GPUs at the Bioinformatics and Computational Biology Center of Colombia (BIOS). Energy was recorded directly via NVML and contrasted with the computational complexity of each model. The results show a Pearson correlation of 0.85 between floating-point operations (GFLOPs) and energy consumption in kWh, confirming that computational complexity is a robust yet imperfect predictor of energy expenditure: architectures with comparable GFLOPs exhibited consumption differing by up to 3.1× due to differences in the hardware efficiency of their dominant operations. The EfficientNet variants offered the best balance between classification performance (Val Top-5 up to 97.15%) and energy efficiency (0.457–0.611 kWh), while Vision Transformers registered the highest relative consumption with the lowest accuracy, evidencing their dependence on large data volumes. These findings guide architecture selection in energy-constrained computing environments.**

## I. Introduction

Deep learning has transformed the field of computer vision, driving remarkable advances in tasks such as image classification, object detection, and semantic segmentation [1]. However, this progress has brought a substantial increase in computational and energy requirements [2]. Training a deep neural network may consume from tenths of a kilowatt-hour to thousands of kWh depending on the architecture, the dataset, and the hardware used [3].

In this context, energy consumption has consolidated as an explicit evaluation criterion in the training of artificial intelligence models. The Green AI agenda proposes treating efficiency as a metric of equal standing to accuracy, reporting the computational and energy cost of developing and training models [4]. Recent studies in this line have further noted that hardware selection directly affects consumption during both training and inference [5], [6]. Meanwhile, the proliferation of modern architectures—convolutional networks, compound-scaling networks, designs derived from neural architecture search (NAS), and vision transformers—offers a wide range of trade-offs between accuracy and computational cost [7].

This work examines seven architectures published between 2019 and 2022, a period in which the field transitioned from traditional ConvNets toward Vision Transformers (ViT), passing through hybrid and systematically scaled designs. These architectures were selected because they represent the state of the art across different recent design paradigms and cover a wide spectrum of computational complexities, from 0.113 to 33.7 GFLOPs—a variation of almost 300× in operations per forward pass—as well as parameter counts between 2.5 and 88.2 million, a difference of more than 34× in model capacity. This diversity makes it possible to analyze how the number of operations relates to real energy consumption during training under identical conditions.

### *A. Evaluated Architectures*

*MobileNetV3 Small and Large.* MobileNetV3, proposed by Howard et al. in 2019 [8], represents the third generation of the MobileNet family, explicitly designed for mobile devices and resource-constrained environments. The architecture combines hardware-aware NAS with the NetAdapt algorithm, an adaptation procedure that automatically adjusts the number of filters in each layer to the latency constraints of the target platform, and further incorporates squeeze-and-excitation attention blocks and the hardware-friendly h-swish activation function. Two variants are included: MobileNetV3-Small, targeting highly constrained resources with 2.54 million parameters and 0.113 GFLOPs, and MobileNetV3-Large, aimed at higher performance under moderate resources, with 5.48 million parameters and 0.433 GFLOPs. Together they cover the lower end of the study's computational efficiency spectrum.

*EfficientNet B0 and B1.* EfficientNet, introduced by Tan and Le in 2019 [9], proposes a compound scaling method that simultaneously balances network depth, width, and resolution through a single coefficient φ. The base architecture, EfficientNet-B0, was designed with NAS and subsequently scaled with this principle to generate variants B1 through B7. EfficientNet-B0 has 5.29 million parameters and 0.772 GFLOPs, while B1 reaches 7.79 million parameters and 1.139 GFLOPs. This approach has been shown to reach superior accuracies with significantly fewer parameters than previous architectures [9], making these networks ideal candidates for evaluating the relationship between operations and energy consumption.

*Vision Transformer ViT-B/32 and ViT-B/16.* Vision Transformers, proposed by Dosovitskiy et al. in 2020 [10], represent a paradigmatic break by applying the Transformer architecture—originally designed for natural language

processing—directly to sequences of image patches, dispensing with convolutional layers. The ViT-B/32 variant divides the image into 32×32 patches, with 8.727 GFLOPs and 88.2 million parameters, while ViT-B/16 uses 16×16 patches, with 33.701 GFLOPs and 86.6 million parameters, which increases the number of tokens and the computational load. Both variants show outstanding performance when pretrained on massive datasets but exhibit low sample efficiency when trained from scratch on medium-sized datasets [11].

*ConvNeXt-Tiny.* ConvNeXt, proposed by Liu et al. in 2022 [12], results from progressively modernizing a standard ResNet by adopting Vision Transformer design principles—such as LayerNorm normalization layers, GELU activations, and 7×7 depthwise convolutions—while maintaining a pure ConvNet structure. The ConvNeXt-Tiny variant, with 28.6 million parameters and 8.911 GFLOPs, is representative of this family and has been shown to compete favorably with Swin Transformers in classification and detection [12]. Its inclusion allows studying how a modern ConvNet design behaves energetically against lightweight models and pure transformers.

The joint selection of these seven architectures allows contrasting, under identical training conditions, three contemporary design paradigms: efficiency-oriented lightweight networks such as MobileNetV3 and EfficientNet, convolutional networks modernized with Transformer principles such as ConvNeXt, and pure vision transformers such as ViT.

### B. Research Problem

Although the Green AI literature has established that the computational cost of deep learning is a sustainability problem [2], [4], most cross-architecture comparisons rely on indirect estimators such as FLOPs or parameter counts, whose reliability as predictors of real energy consumption on specific hardware has been questioned [6], [13]. For the 2019–2022 generation of architectures spanning lightweight networks, compound scaling, modernized ConvNets, and vision transformers, no direct and comparable measurement of training energy consumption exists under strictly uniform conditions of data, hyperparameters, and hardware. The question addressed by this work is: what is the energy consumption of training MobileNetV3-Small, MobileNetV3-Large, EfficientNet-B0, EfficientNet-B1, ViT-B/32, ConvNeXt-Tiny, and ViT-B/16 for ImageNet-1k classification using the same ImageNet-10k subset, under a common 40-epoch baseline configuration on hardware with two P100 GPUs, and to what extent do GFLOPs predict that consumption?

### C. Hypothesis

We posit that, under controlled and uniform training conditions, there is a strong positive association between the computational complexity of an architecture, measured in GFLOPs, and its total energy consumption in kWh, but that this association is not perfect: architectures with comparable GFLOPs may exhibit substantially different energy consumption depending on the hardware efficiency of their dominant operations. Since the experimental design is observational-comparative, with a single run per architecture and a single GPU type, the hypothesis is evaluated through associative evidence—the Pearson correlation coefficient and the analysis of deviations—and not as a causal demonstration generalizable to all hardware.

## II. Related Work

The literature on the energy cost of deep learning can be organized into three lines converging with this study.

The first line establishes the magnitude of the problem and proposes evaluation frameworks. Strubell et al. [2] quantified the energy cost and emissions associated with training language models, showing that architecture search and hyperparameter tuning multiply the expenditure several times. Schwartz et al. [4] formalized the distinction between Red AI, focused exclusively on accuracy, and Green AI, which incorporates efficiency as a first-order evaluation criterion. Bannour et al. [3] reviewed the tools available to estimate the carbon footprint of natural language processing methods, and Lottick et al. [14] proposed energy usage reports as a component of algorithmic accountability. The present work belongs to this agenda but transfers it to the computer vision domain with direct on-hardware measurement.

The second line empirically compares the consumption of different models and configurations. Canziani et al. [6] analyzed convolutional networks in terms of accuracy, memory, and consumption, warning early on that FLOPs are imperfect estimators of real expenditure when implementation efficiency is unknown. Karamchandani and Mozo [5] compared architectures, GPUs, and measurement tools toward an energy consumption index for deep models. Tripp et al. [13] and Fiorini and de Lemos [15] confirmed, with large-scale empirical analyses, that neither FLOPs nor parameters are universally reliable predictors of energy consumed. Regarding training configuration, Cortes et al. [16] reported for ResNet-50 fine-tuning that consumption grows approximately linearly up to 40 epochs but quadruples when moving to 80, while Atiénzar et al. [17] identified learning rate, batch size, numerical precision, and the number of workers as key determinants of energy scalability. These last two works directly ground the methodological decisions of the present experiment.

The third line addresses the relationship between architectural paradigms, data, and hardware efficiency. Maurício et al. [7] systematically reviewed the comparison between Vision Transformers and convolutional networks for classification, and Yuan et al. [11] together with Cao et al. [18] documented the low sample efficiency of ViTs trained from scratch on reduced datasets, attributable to the absence of locality inductive biases. On the hardware side, Yu et al. [19] showed that ConvNeXt-T reaches only about 60% of the training throughput of a ResNet-50 with similar GFLOPs due to the memory-access cost of large-kernel depthwise convolutions, and Qararyah et al. [20] confirmed that depthwise and pointwise convolutions exhibit a lower compute-to-memory-access ratio than standard convolutions, making memory the dominant bottleneck on GPUs.

Unlike these antecedents, which compare model families under heterogeneous conditions or estimate consumption from proxies, this work directly measures the energy expenditure of seven architectures representative of three recent design paradigms under a strictly identical protocol of data, hyperparameters, and hardware, allowing the observed differences to be attributed to the intrinsic properties of each architecture.

## III. Methodology

### A. Computing Infrastructure

All experiments were executed on the infrastructure of the Bioinformatics and Computational Biology Center of Colombia (BIOS), using two NVIDIA Tesla P100 GPUs with 16 GB of HBM2 memory each, connected via NVLink. The P100 belongs to the Pascal high-performance computing family, optimized for deep network training with FP16/FP32 mixed-precision support and a memory bandwidth of 732 GB/s.

### B. Dataset

A stratified subset of ImageNet-1k, called ImageNet-10k, was used, composed of 10,000 images uniformly distributed among the 1,000 classes, i.e., 10 images per class. This subset was used identically for all architectures, guaranteeing direct comparability across experiments.

### C. Baseline Configuration

A uniform training configuration was defined and applied to all models: 40 training epochs, AdamW optimizer with an initial learning rate of $1\times10^{-3}$ and weight decay of $1\times10^{-4}$, a batch size of 64 images per GPU, and standard data augmentation consisting of resizing to 224×224, random horizontal flipping, and normalization with the ImageNet mean and standard deviation.

Adopting a uniform hyperparameter baseline is not a trivial decision from the energy perspective. Atiénzar et al. [17] argue that parameters such as learning rate, batch size, numerical precision, and the number of training workers are key determinants of energy scalability in convolutional networks, as they directly condition convergence dynamics, hardware utilization, and training duration. Fixing a common configuration eliminates this search cost and guarantees that the recorded consumption differences exclusively reflect the architectural properties of each model, not variations in the training configuration.

The choice of 40 epochs as the training horizon is empirically supported by recent literature. Cortes et al. [16] report, in a study of ResNet-50 fine-tuning on the ophthalmological ODIR-5K dataset, that energy consumption scales approximately linearly with the number of epochs up to 40, but that from 80 epochs onward the total consumption nearly quadruples relative to the 40-epoch level, breaking the linear growth pattern. This behavior suggests that 40 epochs constitute a threshold beyond which energy efficiency degrades substantially, justifying its use as the baseline configuration in the present comparative experiment.

### D. Energy Consumption Measurement

The energy consumption of each experiment was recorded via NVML, the power monitoring system integrated in the GPUs of the BIOS computing system, and is reported as the total kilowatt-hours accumulated during the complete execution of the 40 epochs.

### E. Evaluation Metrics

The main metrics recorded for each experiment were: (1) total energy consumption in kWh, (2) training time in seconds, (3) Top-5 validation accuracy on the ImageNet validation set, (4) GFLOPs as a measure of computational complexity per forward pass, and (5) total number of trainable parameters. The Pearson correlation coefficient between GFLOPs and energy consumption was computed to quantify the association between computational complexity and energy expenditure.

## IV. Results

### A. Summary of Results per Architecture

Table I presents the data of each network and the results of the seven training experiments, ordered by run identifier.

TABLE I
Training Results of the Seven Architectures on ImageNet-10k, 40 Epochs, 2× P100 GPUs, BIOS

| Id | Model | GFLOPs | Params | kWh | Time (s) |
|---|---|---|---|---|---|
| 1 | convnext_tiny | 8.911 | 28,589,128 | 2.579 | 38,304 |
| 2 | efficientnet_b0 | 0.772 | 5,288,548 | 0.457 | 7,858 |
| 3 | efficientnet_b1 | 1.139 | 7,794,184 | 0.611 | 10,640 |
| 4 | mobilenet_v3_small | 0.113 | 2,542,856 | 0.137 | 3,847 |
| 5 | mobilenet_v3_large | 0.433 | 5,483,032 | 0.287 | 5,584 |
| 6 | vit_b_16 | 33.701 | 86,567,656 | 3.026 | 32,680 |
| 7 | vit_b_32 | 8.727 | 88,224,232 | 0.833 | 9,935 |

Fig. 1. Scatter plot of GFLOPs vs. energy consumption (kWh) for the seven evaluated architectures. The dashed trend line shows the linear regression ($R^2$ = 0.728). ViT-B/16 and ConvNeXt-Tiny show the largest deviations from the general trend. [Insert figure here.]

### B. Energy Variation Across Architectures

Total energy consumption shows a 22.1× variation between the most efficient architecture and the one with the highest expenditure. MobileNetV3-Small registered the lowest consumption with 0.137 kWh in 3,847 seconds of training, while ViT-B/16 presented the highest consumption with 3.026 kWh in 32,680 seconds. ConvNeXt-Tiny ranked second in energy consumption with 2.579 kWh, although with the longest training time of the experiment, 38,304 seconds, reflecting a lower consumption intensity per unit of time compared to ViT-B/16.

The EfficientNet variants positioned themselves in an intermediate-low consumption range: EfficientNet-B0 consumed 0.457 kWh and EfficientNet-B1 0.611 kWh, representing increases of 3.3× and 4.5× relative to MobileNetV3-Small, respectively. MobileNetV3-Large consumed 0.287 kWh, roughly twice the Small variant, consistent with its higher computational complexity of 0.433 versus 0.113 GFLOPs.

### C. Correlation Between GFLOPs and Energy Consumption

The Pearson correlation analysis between the GFLOPs of each architecture and its energy consumption in kWh yielded a coefficient $r = 0.85$ with $p < 0.05$, indicating a strong and statistically significant positive association. This result suggests that computational complexity measured in floating-point operations is a robust predictor of energy consumption under controlled and uniform training conditions, supporting the first part of the stated hypothesis.

However, the correlation is not perfect, which points to the influence of additional factors and supports the second part of the hypothesis. Notably, ViT-B/32 with 8.727 GFLOPs consumed 0.833 kWh—a relatively low value considering its larger parameter count of 88.2 million—while ConvNeXt-Tiny, with practically identical 8.911 GFLOPs, consumed 2.579 kWh, i.e., 3.1× more. In parallel, ConvNeXt-Tiny consumed almost the same as ViT-B/16 despite having nearly four times fewer GFLOPs, 8.9 versus 33.7. This double discrepancy constitutes one of the most relevant findings of the experiment and requires an explanation from the hardware perspective.

The main cause lies in the nature of the operations dominating each architecture. ConvNeXt-Tiny relies on 7×7 depthwise convolutions, an operation that, despite its low FLOPs, presents an extremely low arithmetic intensity—the ratio between compute and memory access. Yu et al. [19] empirically demonstrate that ConvNeXt-T, with GFLOPs similar to ResNet-50, achieves only 60% of the training throughput on A100 GPUs at full precision, precisely due to the high memory-access costs of large-kernel depthwise convolutions. Qararyah et al. [20] confirm that depthwise and pointwise convolutions have a lower compute-to-memory-access ratio than standard convolutions, turning memory accesses into the dominant performance bottleneck on GPUs. On P100 hardware, whose 732 GB/s memory bandwidth is substantially lower than that of modern GPUs such as the A100, this effect is amplified: the GPU remains stalled waiting for data instead of executing arithmetic operations, which lengthens training time and, with it, total energy consumption. This explains why ConvNeXt-Tiny registered the longest training time of the experiment, 38,304 seconds, exceeding even that of ViT-B/16 at 32,680 seconds.

ViT-B/16, in turn, operates through multi-head self-attention over sequences of image tokens. Although its GFLOP count is 3.8× higher than that of ConvNeXt-Tiny, the attention operations—dense matrix multiplications among the Q, K, and V matrices—are highly parallelizable and well optimized in NVIDIA's cuBLAS/cuDNN libraries, allowing greater utilization of the GPU's compute units. As a result, ViT-B/16 completes its training in less time despite executing considerably more floating-point operations per epoch. The higher instantaneous power of ViT-B/16, derived from greater GPU utilization, combined with a shorter training time, converges toward a total consumption similar to that of ConvNeXt-Tiny, where the GPU operates at lower power but for longer.

This phenomenon illustrates a fundamental principle in the energy efficiency of deep network training: total energy consumed, $E = P \times t$, depends both on instantaneous power and on execution time, and these two factors can offset each other in non-intuitive ways. A model with fewer FLOPs but low hardware efficiency, such as ConvNeXt-Tiny on P100, can consume as much or more energy than one with more FLOPs but greater utilization of compute units, such as ViT-B/16. Canziani et al. [6] already warned that FLOPs are imperfect estimators of real energy consumption when implementation efficiency on the specific hardware is unknown, reinforcing the need for direct measurement—as performed in this work—over estimates based exclusively on the theoretical complexity of the model.

### *D. Relationship Between Accuracy and Energy Efficiency*

From the classification performance perspective, EfficientNet-B0 achieved the highest Top-5 accuracy at 97.15% with the second-lowest energy consumption at 0.457 kWh, establishing itself as the architecture with the best accuracy–energy balance. EfficientNet-B1 followed with 96.91% Top-5 and 0.611 kWh. ConvNeXt-Tiny reached 96.06% with the second-highest energy consumption, 2.579 kWh, while MobileNetV3-Large and MobileNetV3-Small obtained 95.95% and 94.82% with very low consumption.

The ViT variants showed markedly inferior accuracy results: ViT-B/32 reached only 41.49% Top-5 and ViT-B/16 29.65%, despite being the architectures with the largest number of parameters, above 86 million. This behavior is consistent with previous findings showing that Vision Transformers require massive datasets to reach their full potential when trained from scratch [11], [18], exhibiting low sample efficiency on reduced datasets such as the ImageNet-10k employed here.

## V. Discussion

The results confirm the central hypothesis of a strong positive correlation, $r = 0.85$, between GFLOPs and energy consumption in kWh during training. This relationship is consistent with previous studies identifying floating-point operations as the main determinant of computational cost in deep networks [13]. However, the residual variability—the 28% of variance not captured by GFLOPs according to the linear fit in Fig. 1—evidences that internal architecture, parallelism patterns, and GPU implementation efficiency are also determining factors [3], [6]. In terms of the stated hypothesis, the associative evidence supports both the strong association and its imperfect character; generalization to other accelerators remains an open question, since the experiment was executed on a single GPU type.

The disparity between ConvNeXt-Tiny and ViT-B/32, which have comparable GFLOPs, 8.911 versus 8.727, but very different energy consumption, 2.579 versus 0.833 kWh, deserves special attention. The 7×7 depthwise convolutions of ConvNeXt, although computationally equivalent in FLOPs to the attention operations of ViT-B/32, result in lower hardware efficiency on the P100 GPUs used. This finding coincides with what Yu et al. [19] reported regarding the reduced throughput of ConvNeXt-T and with the characterization by Qararyah et al. [20] of the memory bottleneck of depthwise convolutions, and it is consistent with the observations of Tripp et al. [13] and Fiorini and de Lemos [15] that FLOPs and parameters are not universally reliable estimators of energy consumption when the internal architecture is unknown. It should be noted that these antecedents were obtained on more recent hardware; the lower memory bandwidth of the P100 suggests that the penalty observed here represents an upper bound of the effect for later GPU generations.

The low performance of the ViT models on the ImageNet-10k set, 29.65% and 41.49% Top-5 versus CNNs exceeding 94% under the same conditions, highlights one of the best-known limitations of this architecture: its dependence on large volumes of pretraining data [10], [11]. Unlike ConvNets, which incorporate spatial locality inductive biases through the convolution operation, ViTs must learn these structures from the data [18]. With only 10 images per class, this learning capacity proves insufficient. This result coincides with Yuan et al. [11] and Cao et al. [18], who report the need for specific strategies—progressive tokenization or reduced-data regimes with additional supervision—to train vision transformers without massive pretraining.

The decision to fix 40 epochs as the training horizon gains further relevance in light of external evidence. Cortes et al. [16] show that energy consumption grows approximately linearly up to 40 epochs, but that doubling to 80 quadruples the total expenditure, indicating a break in the linear regime that makes 40 epochs an energetically reasonable cutoff point. Additionally, the same authors report that batch size introduces a complex trade-off, as larger batches reduce instantaneous power but may increase total training time and, with it, the energy consumed; in the present work this effect is controlled by fixing a uniform batch size for all architectures.

From an energy-efficiency perspective for constrained computing environments, such as the BIOS context, the results suggest that EfficientNet-B0 and EfficientNet-B1 represent the optimal options for image classification tasks, offering the highest accuracy with moderate consumption. MobileNetV3 is the appropriate choice when minimal consumption is the dominant criterion, while ConvNeXt-Tiny and ViT are more appropriate when greater computing capacity is available or when leveraging pretraining on larger datasets is intended. In this regard, Atiénzar et al. [17] propose treating energy as an explicit scalability metric and integrating energy considerations into hyperparameter optimization, a perspective that reinforces the need to evaluate consumption not only as a function of the architecture but also of the adopted training configurations.

A limitation of the present study is the use of a reduced 10,000-image subset, which inevitably penalizes data-hungry architectures such as ViTs, biasing the accuracy comparison though not the consumption comparison. A second limitation is the execution of a single run per architecture, which prevents estimating variability across repetitions. Future work should replicate the analysis with the full ImageNet-1k dataset of 1.28 million images to evaluate energy behavior under conditions more representative of real practice, as well as on later GPU generations to verify whether the memory penalty of depthwise convolutions is attenuated. Additionally, it would be of interest to explore how pretraining strategies, fine-tuning, and compression techniques such as pruning and quantization modify the relationship between FLOPs and energy consumption.

## VI. Conclusions

This study empirically demonstrated a strong correlation, $r = 0.85$, between computational complexity in GFLOPs and energy consumption in kWh during the training of seven computer vision models on a homogeneous subset of ImageNet-1k. The observed consumption range spans more than 22×, from 0.137 kWh for MobileNetV3-Small to 3.026 kWh for ViT-B/16, evidencing the magnitude of the energy differences among architectural paradigms.

Four main conclusions emerge from the analysis. First, EfficientNet-B0 represents the best balance between accuracy, with 97.15% Top-5, and energy efficiency, with 0.457 kWh, making it the preferred recommendation for energy-constrained computing environments. Second, the Vision Transformers ViT-B/16 and ViT-B/32 show low predictive performance with reduced datasets, reaffirming their dependence on massive pretraining, in addition to presenting high energy consumption relative to the accuracy achieved. Third, GFLOPs are a robust yet imperfect predictor of energy consumption, as factors such as GPU kernel efficiency and memory access patterns explain notable deviations, such as that of ConvNeXt-Tiny versus ViT-B/32 with similar GFLOPs but consumption differing by 3.1×. Fourth, architecture selection in computer vision projects should explicitly incorporate energy consumption as a design criterion, especially in scientific environments with shared resources such as the high-performance computing centers of research institutions.